\documentclass[11pt]{article}
\usepackage[preprint]{acl}
\usepackage{times,latexsym}
\usepackage[T1]{fontenc}
\usepackage[utf8]{inputenc}
\usepackage{microtype,inconsolata}
\usepackage{amsmath,amssymb,booktabs,array,tabularx,graphicx}
\usepackage{algorithm,algpseudocode}
\usepackage{needspace}

\newcommand{\sysname}{DeltaHarness}

\newcommand{\E}{\mathbb{E}}
\newcommand{\Selected}{\textsc{Selected}}
\newcommand{\Uniform}{\textsc{Uniform}}
\newcommand{\Balanced}{\textsc{Balanced}}

\hypersetup{pdftitle={DeltaHarness: Self-Evolving Harnesses for Resource-Adaptive Tool Execution},pdfauthor={Lecheng Yan; Jianze Lin; Xicheng Han; Xiantao Xu; Wenxi Li; Chenyang Lyu}}
\ifdefined\pdfminorversion\pdfminorversion=7\fi
\ifdefined\pdftrailerid\pdftrailerid{}\fi
\title{DeltaHarness: Self-Evolving Harnesses for Resource-Adaptive Tool Execution}
\author{
  Lecheng Yan$^{1}$,
  Jianze Lin$^{2}$,
  Xicheng Han$^{3}$,
  Xiantao Xu$^{4}$, \\
  \bf{Wenxi Li$^{5}$,
  Chenyang Lyu$^{6}$} \\
  $^{1}$University of Science and Technology of China \\
  $^{2}$Beijing Normal University \quad
  $^{3}$Zhejiang University \quad
  $^{4}$Dmall Inc. \\
  $^{5}$East China Normal University \quad
  $^{6}$Alibaba Group \\
  \texttt{lyuchenyang.dcu@gmail.com}
}

\begin{document}
\maketitle
\raggedbottom
\begin{abstract}
Tool-using agents rely on execution harnesses to organize and schedule tool operations. The same task and resource description can conceal runtime conditions that reverse the latency ordering of valid execution plans. Resolving this uncertainty requires firsthand execution in the target setting. We introduce \sysname{}, an execution-guided policy-search method that adapts agent harnesses from cold-start measurements. Its QHI loop uses the observed response to update runtime dispatch, revise physical plans, and select experiments for later proposals. An independent confirmation panel controls policy replacement. With fixed dispatch, a revised materialization policy reduces complete-delivery latency by 30.0\% on one video-editing task across three fresh pairs. On held-out tasks, the evolved policies reduce latency by 8.2\% relative to the best fixed plan in one seed and by 4.6\% relative to the initial proposal in another.
\end{abstract}

\section{Introduction}

Tool-using language models execute tasks through harnesses that organize tool operations and control their execution \citep{lyu2024largelanguagemodelscode,wu2026harnessdev,jiang2026harnessevolve}. Resource-aware execution exposes choices in dataflow structure, input access, and scheduling \citep{poms2018scanner,chaudhry2026murakkab}. A video composition can run as one fused render, as parallel output ranges, or through materialized clips \citep{yan2026crayottertraceablemultiagentworkflows,yan2026crayotterlearninglonghorizonvideo}. Parallel ranges can repeat decoding and add intermediate traffic. Materializing clips spends local compute to change later input access. Source layout, cache state, and interference determine which plan delivers fastest, even across executions with the same visible task and hardware description. \emph{When hidden runtime conditions can reverse plan ordering, identifying the faster plan requires firsthand execution in the target setting.}

Execution-guided search uses measured performance to revise systems algorithms \citep{cheng2025adrs,hamadanian2025glia}, storage configurations \citep{egersdoerfer2026stellar}, and serving policies \citep{jiang2026autopoiesis}. Prior programs and measurements provide an initial cost estimate; the target setting's physical response updates it. Under a fixed measurement budget, adaptation must also choose which comparisons will inform the next revision.

Experiment selection connects harness evolution to task-aware exploration, which values information by its effect on downstream control \citep{wagenmaker2023exploration}. Active video investigation tests competing failure explanations before harness revision \citep{luo2026videorsi}, while self-improving branches adapt the development cases used for search \citep{dong2026branches}. We study whether selecting task--resource measurements can improve a successor's physical execution policy.

\begin{figure*}[t]
\centering
\includegraphics[width=\textwidth]{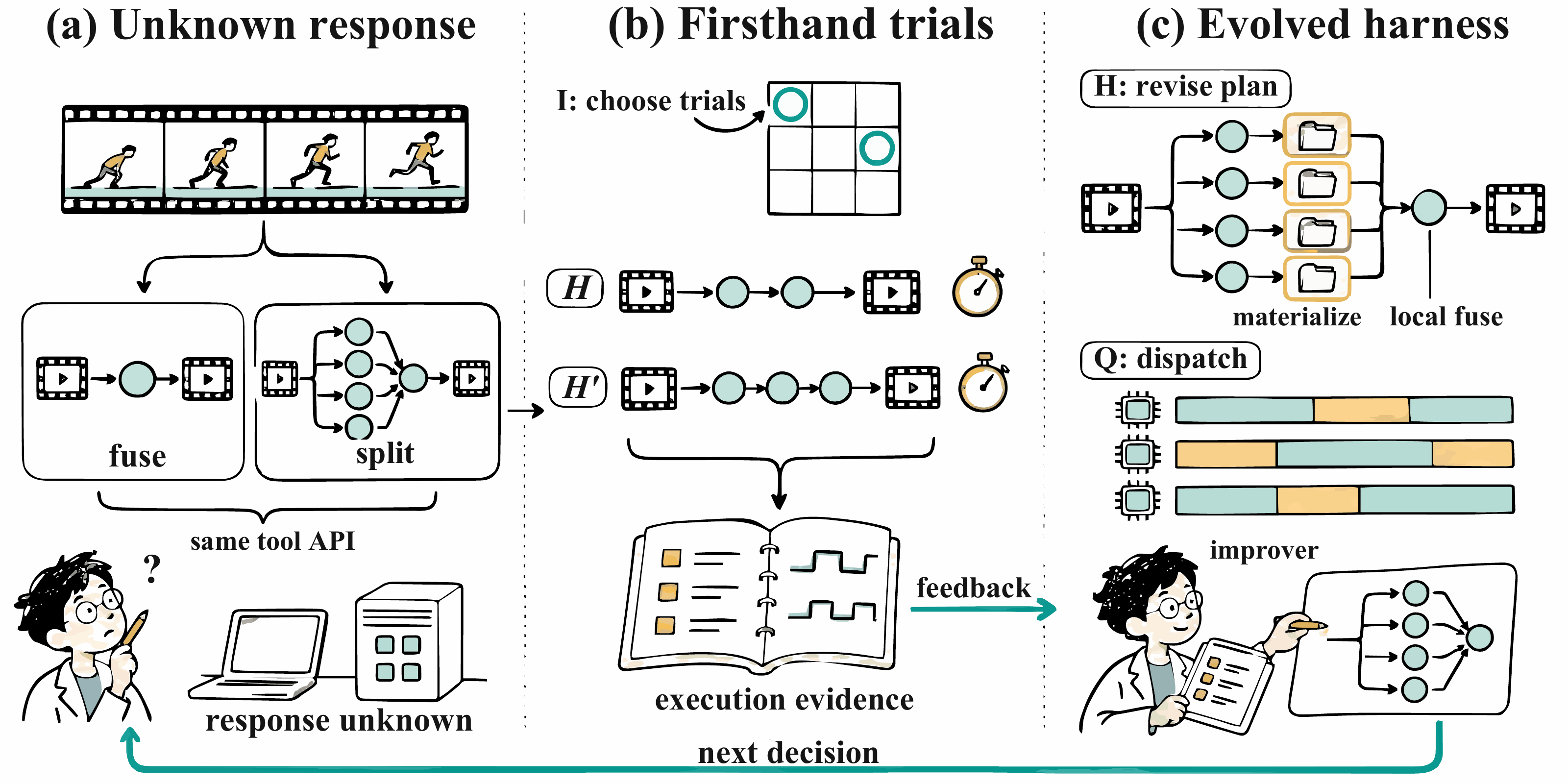}
\caption{Firsthand harness adaptation. (a) A tool interface admits plans with unknown relative cost. (b) $I$ chooses paired trials. (c) The model revises $H$ from the resulting experience; $Q$ dispatches the graph. The illustrated revision materializes clips before local fusion.}
\label{fig:architecture}
\end{figure*}

\sysname{} extends this execution-guided loop to physical tool plans with fixed model weights and tool kernels. For a new device, workload, or resource setting, cold start combines inherited programs and experience with measurements of its current physical response. The physical policy $H$ chooses partitioning, concurrency, fusion, and materialization. The dispatcher $Q$ uses measured delivery times to update admission decisions, and $I$ selects task--resource comparisons for the next proposal. An independent common panel determines incumbent replacement. Figure~\ref{fig:architecture} shows how experience in the target setting informs these revisions.

Program-level recursive self-improvement also makes the improver itself an optimization target \citep{srikanth2026aide2}. Our recursive variant rewrites its evidence-analysis program, while policy self-evolution updates $H$. We compare learned experiment allocation with uniform and balanced panels under matched budgets, followed by frozen held-out evaluation to assess transfer \citep{wang2026rethinking,xia2026rrsi}.

\Needspace{3\baselineskip}
In the fixed-dispatch video-editing case, materializing clips before fusion reduces latency by 30.0\% and raises average active FFmpeg cores from 0.95 to 1.80. On held-out tasks, the first policy-search seed beats the best fixed plan and the second improves its initial proposal. Additional comparisons examine experiment selection, resource response, and orchestration.

\section{Related Work}
\label{sec:related}

\paragraph{Measured Optimization of Computer Systems}

Scanner exposes compressed-video access and dataflow execution \citep{poms2018scanner}, while Murakkab combines workflow profiling with adaptive model and hardware configuration \citep{chaudhry2026murakkab}. LLM-driven systems design builds on measured feedback: ADRS evolves performance-oriented algorithms \citep{cheng2025adrs}; Glia coordinates reasoning, experimentation, and analysis \citep{hamadanian2025glia}; STELLAR reruns applications to tune file systems \citep{egersdoerfer2026stellar}; AUTOPOIESIS evolves serving policies under runtime dynamics \citep{jiang2026autopoiesis}; and MaxKernel uses compiler and profiling feedback for TPU kernels \citep{wang2026maxkernel}. \sysname{} applies this measured-optimization principle to physical tool plans, separating experiment allocation from incumbent confirmation.

\paragraph{Learning and Evolving Agent Harnesses}

HarnessDev evaluates runnable harness creation and evolution \citep{wu2026harnessdev}; HarnessEvolve diagnoses failures through reference trajectories \citep{jiang2026harnessevolve}; and multi-task self-evolution uses the same frozen agent as solver and optimizer \citep{xu2026multitaskharness}. Specialized harnesses accumulate diagnosis experience in OpsHarness \citep{huang2026opsharness}, design feedback in AutoDesign \citep{luo2026autodesign}, and research strategies in ScholarEvolve \citep{yang2026scholarevolve}. Learned editors provide another route: Harness Learning trains a proposer from task feedback \citep{zhang2026harnesslearning}, Harness-R1 trains an engineer through target-agent reruns \citep{shao2026harnessr1}, and Harness-Zero distills harness-induced behavior into solver weights \citep{ye2026harnesszero}. Our formulation fixes the model weights and tool kernels while exposing physical policies and measurement locations to the improver.

\paragraph{Active Evidence and Improving Exploration}

Task-aware experiment design links observation value to its effect on the eventual policy, as in nonlinear system identification for controller optimization \citep{wagenmaker2023exploration}. VideoHarness-RSI evolves video-context constructors \citep{xu2026videoharness}, and Video-RSI actively revisits videos to test failure explanations \citep{luo2026videorsi}. Exploration also improves through historical replay in Dream-RSI \citep{zheng2026dreamrsi}, self-modification records in SelfSearch \citep{yang2026selfsearch}, shared group experience \citep{weng2026gea}, and adaptive development subsets and proposal guidance \citep{dong2026branches}. AIDE2 evaluates rewritten research agents through downstream research performance \citep{srikanth2026aide2}. \sysname{} focuses on fresh candidate--incumbent task/resource pairs: selected measurements inform a successor, while a common panel controls replacement.

\paragraph{Evaluation and Transfer}

RRSI regularizes harness evolution against development-set overfitting \citep{xia2026rrsi}. Rethinking Harness Evaluation compares evolution with task-level search under matched feedback and inference budgets \citep{wang2026rethinking}. AutoTuneBench studies how baseline choice, hardware variation, headroom, and infrastructure affect tuning measurements \citep{chen2026autotunebench}. Controlled agent-scaling experiments further connect architecture benefits to task structure and coordination overhead \citep{kim2025scalingagents}. These findings motivate our fresh paired measurements, frozen held-out evaluation, and separate reporting of execution gains and policy-acquisition cost.

\section{Problem Formulation}
\label{sec:problem}

A task $x$ specifies a logical composition and its delivery contract. The visible resource description $e$ contains quantities such as CPU eligibility or an input-response limit; observations $o$ describe measured behavior in the current setting. A harness policy constructs a physical tool graph
\begin{equation}
G=H(x,e,o),\qquad G=(V,E).
\label{eq:policy}
\end{equation}
Nodes bind fixed tool operations to arguments and resource requests, and edges impose dependencies. The dispatcher $Q$ schedules ready operations. The optimization target is complete-delivery time $T(Q,H;x,e)$, including observation, policy evaluation, compilation, tool execution, and output checks. We write $T(H;x,e)$ when the dispatcher is fixed.

\paragraph{Why firsthand evidence is necessary.}
Let $z$ denote conditions unresolved by $e$, such as cache state or interference. Two plans $p$ and $q$ can share the same visible description yet reverse their expected ordering:
\begin{equation}
\begin{aligned}
\E[T(p)-T(q)\mid x,e,z_1]&<0,\\
\E[T(p)-T(q)\mid x,e,z_2]&>0.
\end{aligned}
\label{eq:unknown}
\end{equation}
When both runtime states are compatible with the available prior, $(x,e)$ leaves the current plan ordering unidentified. Cold-start executions supply observations $o$ of the target setting. These observations update the evidence used for subsequent dispatch, policy revision, and experiment selection.

Let $\mathcal D_r$ be the information available before proposal $r$, $H_0$ the common initial policy, and $H_{r-1}$ the incumbent. A fixed model improver $M$ proposes
\begin{equation}
(\widetilde H_r,\widetilde A_r)
=M(H_0,H_{r-1},\mathcal D_r),
\label{eq:proposal}
\end{equation}
where $\widetilde A_r$ names candidate--incumbent comparisons on development task--environment cells. The measurement rule $I$ determines which locations are executed. We compare measurement rules through the final policy they produce under the same proposal and execution budgets.

Policy acquisition has a one-time cost $C_{\mathrm{improve}}$. Over $n$ deployed tasks, its time cost is
\begin{equation}
C_{\mathrm{improve}}+\sum_{j=1}^{n}T(Q,H;x_j,e_j).
\label{eq:total}
\end{equation}
Held-out execution ratios assess the reusable policy; model calls, development measurements, failures, and waits determine the acquisition term.

\begin{figure*}[t]
\centering
\includegraphics[width=\textwidth]{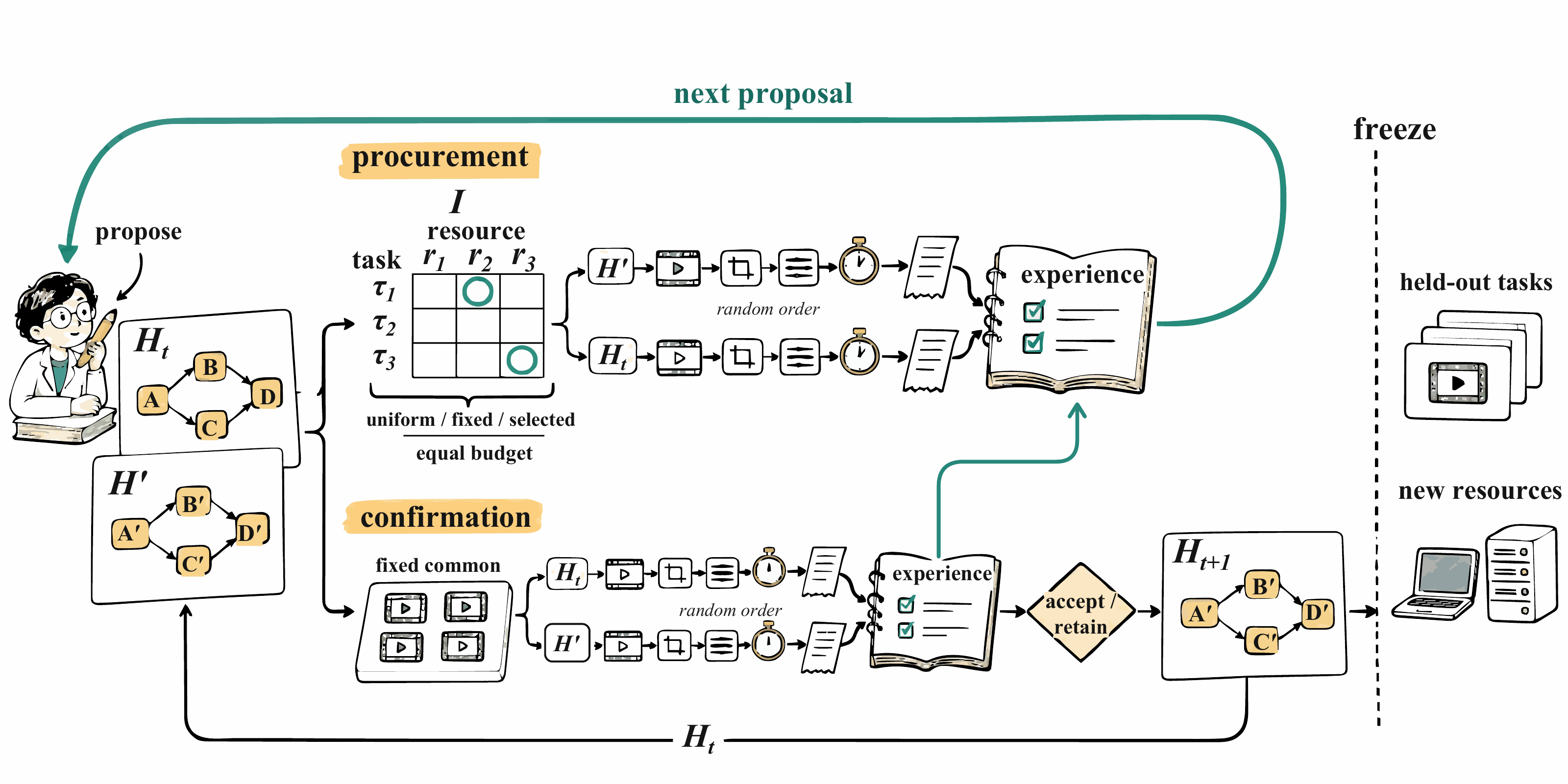}
\caption{The update loop in Algorithm~\ref{alg:deltaharness}. Selected trials supply feedback; common confirmation controls incumbent replacement. Both panels use fresh randomized-order pairs. Frozen policies are evaluated on held-out tasks and resources.}
\label{fig:learning_loop}
\end{figure*}

\section{DeltaHarness}
\label{sec:architecture}

\sysname{} performs black-box policy optimization with adaptive experiment allocation. In the full loop, $H$ constructs the physical graph and $Q$ decides whether to admit a ready operation or wait for an active one to finish. Successful workflows supply decision-to-delivery times for updating $Q$. The improver proposes revisions to $H$ and $I$ from accumulated outcomes. The physical-policy and allocation experiments use fixed dispatch.

\subsection{Representing Physical Policies}

A policy maps task and resource observations to a complete physical plan. The Linux studies implement this map as sandboxed source programs. The Mac adaptation case uses ordered typed rules with a complete default branch: conditions test task or environment features and define missing-value behavior. These are distinct policy interfaces over the same kinds of physical choices. The compiler and tool kernels are fixed within each comparison.

For video editing, the plan chooses range count, range concurrency, even or clip-aware cuts, longest-processing-time ordering, hardware decoding, clip materialization, materialization concurrency, and audio/video insertion order (Appendix~\ref{app:q}). These choices change the physical work. Materialization creates local inputs for a fused render (Figure~\ref{fig:architecture}(c)); scheduling controls when the resulting nodes execute.

\subsection{Obtaining Firsthand Measurements}

Cold start acquires response measurements from the current setting and combines them with inherited evidence. An observation includes elapsed time, tool CPU time, input service, work size, and delivery status. Task preparation supplies observations and reusable artifacts, with its cost included in each method's execution. Additional plan comparisons provide evidence for later decisions.

A comparison cell $u=(x,e)$ binds a candidate and incumbent to the same task and environment. Each policy executes freshly, consecutively, in randomized order. Even identical plans receive separate executions. The resulting record contains the common task observation, both policies, both delivery outcomes, and both costs. Remeasuring the incumbent controls for drift better than comparison with a stale timing.

Admission limits constrain declared tool pools; FFmpeg retains natural internal threading. Linux CPU-affinity and FUSE-response controls, and Mac admission and HTTP-response controls, retain their respective resource semantics throughout evaluation.

\subsection{Selecting Evidence for the Next Revision}
\label{sec:learning}

Algorithm~\ref{alg:deltaharness} describes policy evolution with fixed dispatch. Selected trials $A_r$ inform later proposals. Confirmation $C_r$ determines whether the current candidate replaces the incumbent. Both panels contribute to the accumulated experience in Figure~\ref{fig:learning_loop}. The allocation experiment uses source policies.

\begin{algorithm}[t]
\caption{Policy evolution with fixed dispatch}
\label{alg:deltaharness}
\small
\begin{algorithmic}[1]
\Require Common policy $H_0$, prior $\mathcal D_1$, model $M$, measurement rule $I$
\Require Development domain $\mathcal U$, fixed panel budgets $b^A_r,b^C_r$ for $R$ rounds
\For{$r=1,\ldots,R$}
  \State $(\widetilde H_r,\widetilde A_r)\gets M(H_0,H_{r-1},\mathcal D_r)$
  \State Freeze candidate $\widetilde H_r$ and locations $\widetilde A_r$
  \State $A_r\gets I(\widetilde A_r,\mathcal D_r,\mathcal U,b^A_r)$
  \State Reveal common panel $C_r$ of size $b^C_r$
  \State $F^C_r\gets\Call{Pair}{\widetilde H_r,H_{r-1},C_r}$
  \State $F^A_r\gets\Call{Pair}{\widetilde H_r,H_{r-1},A_r}$
  \State $H_r\gets\Call{RetainOrReplace}{H_{r-1},\widetilde H_r,F^C_r}$
  \State $\mathcal D_{r+1}\gets\mathcal D_r\cup\{F^C_r,F^A_r,H_r\}$
\EndFor
\State Freeze $H_R$ for held-out evaluation
\end{algorithmic}
\end{algorithm}

\paragraph{Measurement allocation.}
In the allocation study, the three LLM arms share model, prompt structure, sampling settings, and proposal count. \Uniform{} samples development cells without replacement within a panel. \Balanced{} uses a predefined balanced panel. \Selected{} executes the model's locations, including repeats. Invalid locations receive uniform replacement. The model packet omits arm labels. A $(1+1)$-ES baseline mutates resource-bin plan parameters under the same execution budget, without additional LLM calls.

All arms begin from one $H_0$ generated before iterative candidate--incumbent feedback. They share the prior and pilot, then receive their own measurements. Each subsequent input retains the initial policy, incumbent, previous candidates, queries, acceptance decisions, failures, and costs. Pair records contain both outcomes and coordinates.

\paragraph{Incumbent replacement.}
The common panel is revealed after candidate freezing. For paired losses $\ell'_{rj}$ and $\ell_{rj}$ on $C_r$, define
\begin{equation}
\rho_r=\exp\left(\frac{1}{|C_r|}
\sum_{j\in C_r}\log\frac{\ell'_{rj}}{\ell_{rj}}\right).
\label{eq:accept}
\end{equation}
The plan-search and allocation studies use delivery time as loss for complete pairs and predefined penalties for failed deliveries. Replacement requires $\rho_r\leq0.995$ and a one-sided 90\% upper bound below 1 from 2,000 cell-bootstrap replicates. In Algorithm~\ref{alg:deltaharness}, only $C_r$ enters this decision. Every candidate receives its allocated measurements, including slower candidates.

\paragraph{Feedback and cost.}
The eight-round allocation study uses 24 confirmation and 24 selected pairs per round, totaling 768 executions per arm and 25 LLM proposals including the common initial proposal. Each proposal receives outcomes from completed previous rounds. The acquisition cost includes the final round's measurements.

After development, policies are frozen and evaluated once on new tasks and resource conditions. The intended allocation comparisons are \Selected{}/\Uniform{} and \Selected{}/$H_0$, with task-clustered uncertainty and complete delivery.

\paragraph{Recursive variant.}
Plan search evaluates candidate--incumbent pairs on the full 96-cell development panel each round. Its recursive variant changes the improver's \texttt{analyze(data)} program and method notes every two rounds. That program summarizes observations for subsequent proposals. Comparison with a policy-only improver measures the additional recursive layer (Appendix~\ref{app:q}).

\section{Evaluation}
\label{sec:evaluation}

We evaluate firsthand policy revision, held-out policy search, recursive analysis rewriting, and experiment allocation under fixed dispatch. Each matched comparison fixes the task, resource condition, and tool implementation. Appendix~\ref{app:historical} gives additional server execution, orchestration, and scheduling measurements.

\paragraph{Firsthand adaptation case.}
One fixed 312-frame task runs on a 12-core Mac with stage-pool cap four and shared HTTP input-response shaping at 4~MiB/s. The valid reference $H_P$ uses four clip-aware ranges with LPT ordering. The configured model, \texttt{gpt-6.1-terra}, receives accumulated execution history, compares three alternatives, and revises its policy. Three fresh randomized-order pairs evaluate the frozen revision under the same source bytes, compiler, and resource condition. The predefined development budget allows up to three such rounds; this case finishes in the first round.

\paragraph{Frozen-policy transfer.}
We freeze the Mac policy and reference before evaluating eight predefined, previously unexecuted edit configurations drawn from known media. Stage cap four crosses local input and shared HTTP response limits of 4 and 1.5\,MiB/s, with two fresh randomized-order pairs per task and condition: 48 pairs and 96 executions. The primary panel is 4\,MiB/s. The two policies choose the same physical plan on local input, providing a control for measurement variability. Frame completeness, latency, video fidelity, and audio duration are reported separately.

\paragraph{Plan evolution.}
The plan-search study develops policies on six tasks and sixteen environments: CPU-affinity sizes 2/4/8/16 crossed with local input or FUSE rates 16/4/2~MiB/s. Eight held-out tasks cross 24 environments, adding 8/3~MiB/s, for 192 cells per seed. Qwen3.8-27B generates ten candidates for a \emph{policy-only improver} with a fixed analysis procedure or a \emph{recursive improver} that also rewrites its analysis program and method notes. Table-based ES uses the same paired evaluator. The first policy-only proposal is retained for a secondary no-feedback comparison. Two seeds run under different observed CPU contention.

\paragraph{Measurement selection.}
The Linux allocation study uses eight search and four development-validation tasks over sixteen environments. \Uniform{}, \Balanced{}, \Selected{}, and ES each receive eight rounds with 24 confirmation and 24 selected pairs. LLM arms share Qwen3.8-27B, temperature 0.6, and a 16,000-token generation limit. Final evaluation uses sixteen new edit configurations over known media and 24 environments, including new 6/1.5~MiB/s conditions. This experiment uses source policies; it does not evaluate ES over the Mac typed-rule interface.

\paragraph{Fixed plans and inference.}
Range4, range8, K2p2, and fused form the fixed plan bank; K2p2 uses two ranges with pool two. The bank oracle takes the fastest measured plan per cell. Ratios below 1 favor the numerator. Held-out intervals use 10,000 task-cluster bootstrap replicates, keeping a task's environments together. Delivery failures are counted separately from physical latency; predefined penalties are used only for search losses. The two plan-search seeds are analyzed within their respective host conditions.

\paragraph{Resource and delivery checks.}
FFmpeg retains natural internal threading. Affinity controls CPU eligibility, and admission controls declared pools, on shared hosts. The executor checks output properties; full AV decoding follows timing. The retained quality results and contention-retry protocol are in Appendix~\ref{app:completed}. Quality coverage includes format, decodability, and selected source/output comparisons.

\section{Results}
\label{sec:results}

\subsection{Firsthand Revision Improves a Valid Plan}
\label{sec:performancecase}

Firsthand comparisons identify clip materialization followed by local fusion as an effective alternative to the four-range reference $H_P$ under constrained input service. The revised policy uses a materialization pool of four in this setting. Table~\ref{tab:performancecase} gives the fresh confirmation results, and Appendix~\ref{app:historical} reports the three alternatives that informed the revision.

\begin{table}[ht]
\centering
\setlength{\tabcolsep}{5pt}
\renewcommand{\arraystretch}{1.08}
\begin{tabular}{@{}ccrrr@{}}
\toprule
Trial & Order & $H$ (s) & $H_P$ (s) & Ratio \\
\midrule
1 & $H\to H_P$ & 9.74 & 14.06 & 0.693 \\
2 & $H\to H_P$ & 9.75 & 14.27 & 0.683 \\
3 & $H_P\to H$ & 9.73 & 13.42 & 0.725 \\
\midrule
\multicolumn{4}{@{}l}{\textit{Paired geometric mean}} & \textbf{0.700} \\
\bottomrule
\end{tabular}
\caption{Three fresh paired confirmations on the fixed task. Both policies deliver all 312 frames with audio; $H/H_P$ is the complete-delivery latency ratio.}
\label{tab:performancecase}
\end{table}

The updated policy wins all three pairs and reduces complete-delivery latency by 30.0\%. Average active FFmpeg cores, measured as tool CPU seconds divided by delivery time, rise from 0.95 to 1.80. Cumulative FFmpeg CPU seconds increase by 32.4\%, while source-response payload falls from 44.7--46.9\,MB to 19.9--20.1\,MB. Materialization mobilizes local compute and approximately halves traffic on the constrained input path.

\subsection{Transfer, Correctness, and Video Fidelity}
\label{sec:frozen_transfer}

\begin{figure}[t]
\centering
\includegraphics[width=\columnwidth]{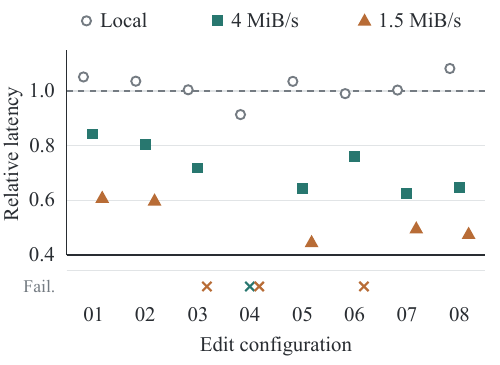}
\caption{Frozen-policy transfer: per-configuration $H/H_P$ latency, geometric mean of two fresh pairs. Values below one favor $H$. Crosses in the lower strip mark reference frame deficits, which leave the full-panel speed ratio undefined. All $H$ outputs complete.}
\label{fig:transfer_response}
\end{figure}

The frozen-policy evaluation reveals both latency changes and execution failures (Figure~\ref{fig:transfer_response}). The materialization policy completes 48/48 executions; the reference completes 40/48. Eight reference executions lose frames under input shaping, leaving the all-task primary speed comparison undefined. These incomplete outputs are recorded as execution failures.

Conditional on both outputs being complete, latency reductions range from 15.7--37.5\% at 4\,MiB/s and 39.1--55.3\% at 1.5\,MiB/s. Local input provides a same-action control: per-configuration ratios span 0.914--1.082, with an aggregate ratio of 1.014 [0.979, 1.044]. This spread shows the measurement variability with two pairs per configuration.

Materialization adds intermediate re-encoding. On fully aligned shaped-input pairs, SSIM is 0.9843--0.9956 and PSNR is 32.8--46.4\,dB, so the latency reductions accompany a measurable fidelity loss. Audio sample counts match in all pairs. Appendix~\ref{app:completed} gives the delivery and fidelity breakdown.

\subsection{Policy Evolution Across Two Seeds}

\begin{figure}[t]
\centering
\includegraphics[width=\columnwidth]{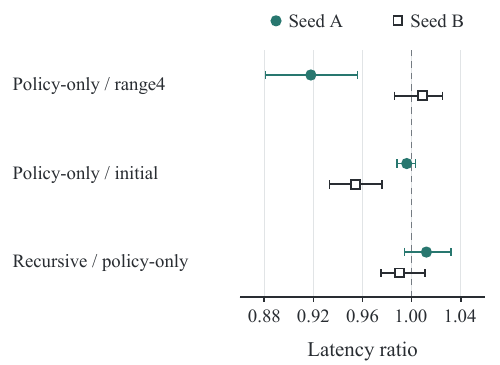}
\caption{Within-seed latency ratios and task-clustered 95\% intervals on eight held-out tasks and 192 cells per seed. Values below one favor the numerator. Seed B has higher external CPU contention. Recursive denotes analysis rewriting; initial denotes the first policy-only proposal.}
\label{fig:evolution_comparison}
\end{figure}

The first plan-search seed obtains an 8.2\% reduction relative to range4, the best fixed plan in that run, and a 6.6\% reduction on unseen input-response conditions. The evolved/initial ratio is 0.996 [0.988, 1.003] (Figure~\ref{fig:evolution_comparison}). The evolved and initial policies close 73.0\% and 69.0\% of the logarithmic bank-oracle gap. The policies select fused execution at low input-service rates and use task features to condition the physical plan.

The second seed runs under higher external CPU occupancy, approximately 0.69 versus 0.01 on assigned cores. Policy-only iteration reduces latency relative to its initial proposal by 4.6\%, with ratio 0.954 [0.933, 0.976]. Figure~\ref{fig:evolution_comparison} and Table~\ref{tab:planresults} report the policy and analysis-rewriting comparisons in both seeds. Table~\ref{tab:fixedbank} gives all fixed-plan references.

\subsection{Physical Response and Server Execution}
\label{sec:platforms}

\begin{figure}[t]
\centering
\includegraphics[width=\columnwidth]{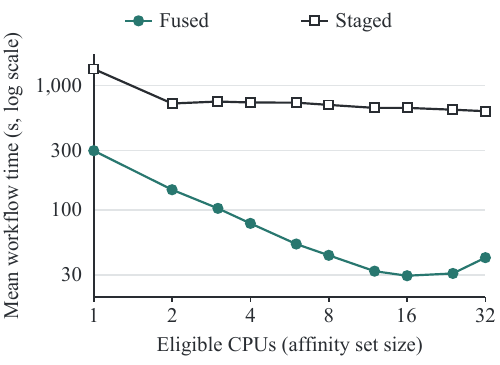}
\caption{Physical-plan response to CPU affinity. Each point averages six compositions, with one observation per composition, plan, and setting. FFmpeg uses automatic threads under a 128\,MiB/s logical read/write service. Affinity-set size specifies eligible CPUs on a shared host.}
\label{fig:cpu_response}
\end{figure}

Controlled CPU-affinity measurements reveal different resource responses for the physical plans (Figure~\ref{fig:cpu_response}). Fused execution benefits from larger affinity sets beyond the point where staged execution largely flattens. This plan-dependent response motivates conditioning the harness on both task structure and available resources.

Additional within-host comparisons cover servers equipped with A100, RTX~3090, and RTX~4090 GPUs. A DAG-aware rule shortens six-task episodes relative to native concurrency-three execution in five host/resource conditions (Appendix~\ref{app:historical}). That appendix also separates a model-call orchestration gain from native rendering time and reports a same-program feedback comparison.

\subsection{Measurement Allocation}
\label{sec:allocation_results}

Table~\ref{tab:selection} reports the final policies from uniform, balanced, and model-selected measurement schedules on sixteen held-out tasks and 384 cells. Each schedule receives the same eight-round proposal and measurement budget. Appendix~\ref{app:completed} gives proposal-validity counts, successful-feedback timing, fixed-plan comparisons, and results by resource condition.

\begin{table}[ht]
\centering
\setlength{\tabcolsep}{6pt}
\renewcommand{\arraystretch}{1.08}
\begin{tabular}{@{}lrr@{}}
\toprule
Comparison & Ratio & 95\% CI \\
\midrule
S / Uniform & 0.984 & [0.966, 1.001] \\
S / Balanced & 1.173 & [1.103, 1.251] \\
\bottomrule
\end{tabular}
\caption{Final allocation-study outcomes on 16 held-out tasks and 384 cells, with task-clustered intervals. S denotes Selected. All three learned policies complete the evaluation; the shared initial policy and ES fail on all 384 cells.}
\label{tab:selection}
\end{table}

\Needspace{9\baselineskip}
\section{Discussion}
\label{sec:discussion}

Cold-start measurements contribute through the execution choices they change. In the adaptation case, firsthand comparisons lead the improver to materialize clips before fusion. This reduces source traffic and exposes more local concurrency, lowering delivery latency as CPU utilization rises. The transfer measurements also reveal the fidelity effects of intermediate re-encoding. The scheduling comparison quantifies the interaction between useful overlap and interference with active work. These outcomes illustrate how observations from the target setting guide physical harness decisions.

\Needspace{9\baselineskip}
\section{Conclusion}

\sysname{} uses cold-start measurements to learn how execution plans behave under the current task and resource conditions. This information guides dispatch, physical-policy revision, and further measurement. With fixed dispatch and tool kernels, a model-selected materialization plan reduces complete-delivery latency by 30.0\% on one task. Held-out policy search beats the best fixed plan in one seed and improves its initial policy in another.

\bibliography{references}
\bigskip
\appendix

\section{Policy Interfaces and Recursive Variant}
\label{app:q}

\paragraph{Physical policy interface.}
The Mac policy evaluates ordered conditions, assigns missing-value behavior, and ends with a complete default plan. Task features include source intervals, clip count, output frames, transitions, and estimated decode work. The Linux studies use sandboxed source policies. Table~\ref{tab:plans} lists the physical choices exposed by the executor; each comparison fixes its compiler and tool implementation.

\begin{table}[ht]
\centering
\begin{tabularx}{\columnwidth}{@{}lX@{}}
\toprule
Parameter & Physical choice \\
\midrule
$K$ & Number of output ranges, 1--32 \\
Range pool & Concurrent range operations \\
Cut mode & Even or clip-aware cuts \\
LPT & Order ranges by estimated work \\
Hardware decode & Request when supported \\
Materialize & Create intermediate clips \\
Materialization pool & Concurrent intermediates \\
Audio first & Audio/video insertion order \\
\bottomrule
\end{tabularx}
\caption{Typed physical policy parameters. CPU-only experiments fix hardware decoding off.}
\label{tab:plans}
\end{table}

\paragraph{Recursive analysis rewriting.}
The policy-only improver has a fixed method description and analysis procedure. The recursive improver rewrites both its method description and a sandboxed \texttt{analyze(data)} function every two rounds. This function reads the arm's accumulated task features, resource observations, plan parameters, wall times, tool CPU times, and I/O counts. Its output enters later policy prompts. If $a_r$ is that analysis program, the additional update is
\begin{equation}
\begin{aligned}
a_{r+1}&=\mathcal R(a_r,\mathcal D_r),\\
\widetilde H_{r+1}&=M(H_r,\mathcal D_r,a_{r+1}(\mathcal D_r)).
\end{aligned}
\end{equation}
Model weights and the experiment-selection procedure remain fixed in this analysis-rewriting comparison. Rewriting the experiment-selection program $I$ would constitute another recursive layer.

\section{Evaluation Details}
\label{app:completed}

\paragraph{Two-seed plan search.}
The policy-only and recursive arms share Qwen3.8-27B, the paired evaluator, and a 32,000-token generation limit. If generation exhausts its budget before returning code, a direct-answer continuation is used; 32 of 56 first-seed calls use this rule. Both arms make ten candidate proposals. The no-feedback comparator is the first policy-only proposal. Table~\ref{tab:planresults} gives the three LLM policies, and Table~\ref{tab:fixedbank} gives the fixed-plan references. Recursive/policy-only ratios are 1.012 [0.994, 1.032] in seed A and 0.990 [0.975, 1.011] in seed B. ES has overall ratio 1.006 [1.003, 1.009] in seed A and point estimate 1.004 in seed B. Its unseen-condition ratios are 1.012 and 1.003. Output checks include format validation and sampled SSIM.

\begin{table*}[t]
\centering
\setlength{\tabcolsep}{4pt}
\begin{tabular}{@{}lrrrrrr@{}}
\toprule
& \multicolumn{3}{c}{Seed A} & \multicolumn{3}{c}{Seed B} \\
\cmidrule(lr){2-4}\cmidrule(l){5-7}
Policy & Ratio & 95\% CI & Unseen & Ratio & 95\% CI & Unseen \\
\midrule
Policy-only & 0.918 & [0.881, 0.956] & 0.934 & 1.009 & [0.986, 1.025] & 1.076 \\
Initial proposal & 0.922 & [0.878, 0.967] & 0.939 & 1.058 & [1.020, 1.093] & 1.107 \\
Recursive analysis & 0.929 & [0.897, 0.958] & 0.958 & 0.999 & [0.990, 1.009] & 1.003 \\
\bottomrule
\end{tabular}
\caption{Plan-search latency relative to range4, eight held-out tasks and 192 cells per seed. Unseen input-response conditions are 8 and 3\,MiB/s. Intervals are task-clustered. External CPU occupancy differs between seeds.}
\label{tab:planresults}
\end{table*}

\begin{table}[ht]
\centering
\setlength{\tabcolsep}{7pt}
\begin{tabular}{@{}lrr@{}}
\toprule
Fixed plan & Seed A & Seed B \\
\midrule
Range4 & 1.000 & 1.000 \\
Range8 & 1.030 & 0.967 \\
K2p2 & 1.020 & 1.091 \\
Fused & 1.070 & 1.210 \\
\midrule
Bank oracle & 0.889 & 0.848 \\
\bottomrule
\end{tabular}
\caption{Fixed-plan point estimates and per-cell bank oracle relative to range4. The best fixed plan changes from range4 to range8.}
\label{tab:fixedbank}
\end{table}

\paragraph{Allocation-study validity.}
Eight search tasks and four development-validation tasks are selected before execution. Final evaluation uses sixteen new complete edit configurations; a separate 32-cell development-validation diagnostic receives no later model feedback. The shared 64-cell pilot has best-fixed geometric mean 32.83\,s and bank-oracle mean 27.33\,s, a 16.7\% gap. Every arm receives the complete pilot. A descriptive best-plan-per-environment combination closes 91.8\% of the logarithmic pilot gap.

The common initial source violates its execution contract and is used without a successful initial-policy confirmation. ES mutations inherit the invalid source; several LLM proposals also fail the response contract. Successful development executions are Uniform 624/768, Balanced 336/768, Selected 48/768, and ES 0/768. Of Selected's 192 requested locations, 120 are model choices and 72 use uniform replacement. Its successful executions follow proposal eight. Uniform, Balanced, and Selected complete all 384 final cells; $H_0$ and ES fail on all 384. Their latency ratios to range4 are 1.292 for Uniform, 1.084 for Balanced, and 1.272 for Selected. The Selected/$H_0$ ratio of 0.636 [0.590, 0.686] is a failure-penalty statistic. Selected's logarithmic bank-oracle gap closure is $-120.8\%$ [$-161.7\%$, $-89.8\%$]. Table~\ref{tab:environments} retains the resource-level outcomes.

\begin{table}[ht]
\centering
\setlength{\tabcolsep}{7pt}
\renewcommand{\arraystretch}{1.08}
\begin{tabular}{@{}lrr@{}}
\toprule
Response (MiB/s) & S / Uniform & S / Balanced \\
\midrule
Local & 0.985 & 0.991 \\
16 & 0.947 & 1.002 \\
6 (unseen) & 1.000 & 1.683 \\
4 & 0.971 & 1.516 \\
2 & 1.005 & 1.008 \\
1.5 (unseen) & 0.998 & 1.020 \\
\bottomrule
\end{tabular}
\caption{Measurement-selection results by response condition, 64 cells per row. S denotes Selected. Its largest losses to Balanced occur at 6 and 4~MiB/s.}
\label{tab:environments}
\end{table}

\paragraph{Mac local-I/O characterization.}
This pilot uses \texttt{deepseek-v4.1-flash}, distinct from the model in the Mac adaptation case. Four tasks cross stage caps 1/2/4/8. Fixed-plan geometric means are 4.60\,s for range4, 5.02\,s for range8, 4.73\,s for K2p2, and 6.15\,s for fused; the oracle is 4.50\,s. The 2.1\% gap is below the predefined 5\% study premise, so this pilot proceeds to no policy-evolution or held-out evaluation.

The contention rule permits two extra whole-block attempts above 10\% interference. It retains the first complete attempt with known low contention, or the third attempt with its flag. Unknown contention remains flagged. This produces 180 pilot executions and thirty successful workflow-check executions; actual costs include all attempts.

\paragraph{Transfer outcomes and fidelity.}
Tables~\ref{tab:frozen_transfer} and~\ref{tab:transfer_tasks} report every condition and configuration. Each ratio uses two pairs with complete video on both sides. Reference frame deficits occur on configurations 03, 04, and 06 under input shaping. All local deliveries complete.

\begin{table}[ht]
\centering
\setlength{\tabcolsep}{5pt}
\begin{tabular}{@{}lrr@{}}
\toprule
Input & $H$ & $H_P$ \\
\midrule
Local & 16/16 & 16/16 \\
4\,MiB/s & 16/16 & 14/16 \\
1.5\,MiB/s & 16/16 & 10/16 \\
\bottomrule
\end{tabular}
\caption{Complete frozen-policy deliveries. Each condition contains eight new configurations and two pairs per configuration.}
\label{tab:frozen_transfer}
\end{table}

\begin{table}[ht]
\centering
\setlength{\tabcolsep}{5pt}
\begin{tabular}{@{}lrrr@{}}
\toprule
Task & Local & 4\,MiB/s & 1.5\,MiB/s \\
\midrule
01 & 1.051 & 0.843 & 0.609 \\
02 & 1.036 & 0.805 & 0.599 \\
03 & 1.005 & 0.718 & Fail \\
04 & 0.914 & Fail & Fail \\
05 & 1.035 & 0.645 & 0.447 \\
06 & 0.991 & 0.761 & Fail \\
07 & 1.004 & 0.625 & 0.497 \\
08 & 1.082 & 0.648 & 0.477 \\
\bottomrule
\end{tabular}
\caption{Per-task frozen $H/H_P$ latency ratios. Every $H$ delivery completes. Fail denotes a reference frame deficit in both repetitions, leaving the speed ratio undefined.}
\label{tab:transfer_tasks}
\end{table}

All 96 outputs decode without decoder errors. Frame counts confirm 48 complete $H$ outputs and 40 complete $H_P$ outputs; paired audio sample counts match throughout. Of the 40 complete video pairs, 32 have full-length SSIM/PSNR records; eight metric records end one frame before the decoded output. Table~\ref{tab:transfer_quality} reports the fully aligned subset. The resulting fidelity loss reflects intermediate re-encoding.

\begin{table*}[t]
\centering
\setlength{\tabcolsep}{7pt}
\begin{tabular}{@{}lrrr@{}}
\toprule
Input & Full metric pairs / 16 & SSIM range & PSNR range (dB) \\
\midrule
Local & 16/16 & 1.0000 & $\infty$ \\
4\,MiB/s & 10/16 & 0.9843--0.9956 & 32.8--46.4 \\
1.5\,MiB/s & 6/16 & 0.9843--0.9956 & 40.5--46.4 \\
\bottomrule
\end{tabular}
\caption{Video fidelity relative to $H_P$ on fully frame-aligned transfer pairs. Eight pairs have reference frame deficits; eight additional complete-output pairs have a one-frame-short metric record. Intermediate re-encoding produces measurable image distortion.}
\label{tab:transfer_quality}
\end{table*}

\section{Additional Execution Studies}
\label{app:historical}

\paragraph{Server execution.}
Table~\ref{tab:platformhistory} compares two fixed methods on three server stacks. Native admits ready tool operations through a concurrency-three pool; DAG rule is a graph-dependent physical-plan selector using the revealed dependencies. Five paired blocks contain six-task episodes under each resource sequence. The machine-readable supplement retains the original eight-policy comparison; the table here addresses the defined native-versus-DAG comparison.

\begin{table*}[t]
\centering
\setlength{\tabcolsep}{6pt}
\begin{tabular}{@{}lrrrrr@{}}
\toprule
& A100 host & RTX3090 host & \multicolumn{3}{c}{RTX4090 host} \\
\cmidrule(lr){2-2}\cmidrule(lr){3-3}\cmidrule(l){4-6}
Policy & CPU drift & CPU drift & Stable & CPU drift & Export drift \\
\midrule
Native concurrency three & 543.25 & 268.55 & 284.64 & 325.80 & 323.66 \\
DAG rule & 370.92 & 236.92 & 235.85 & 293.76 & 282.91 \\
\bottomrule
\end{tabular}
\caption{Server execution comparison: mean seconds per six-task episode, five paired blocks per condition. Native uses a concurrency-three pool; DAG rule is a fixed graph-dependent plan selector. Accelerator names identify hosts. Episode times include execution and diagnostics.}
\label{tab:platformhistory}
\end{table*}

\paragraph{Controlled CPU response.}
Figure~\ref{fig:cpu_response} covers six compositions at ten CPU-affinity sizes: 120 executions, one observation per case, plan, and setting, in fixed pair order. Every execution completes. A private 128\,MiB/s logical read/write service applies to each unit while host resources remain shared. The curves plot workflow time, with setup and diagnostics accounted for separately.

\paragraph{Model-authored orchestration.}
A render-first harness executes a known composition without routine model round trips. Four development cases match two CPU-affinity settings and two forced physical plans (Table~\ref{tab:renderfirst}). Worker time falls by 29.8\%, with 118.0\,s of normal generation calls removed. Native rendering time is 275.73\,s before and 277.53\,s after the change. This gain concerns the model-call path. Revised outputs satisfy the reported twelve-second, 1080p, 30-fps, audio-present checks and decode successfully.

\begin{table}[ht]
\centering
\setlength{\tabcolsep}{5pt}
\begin{tabular}{@{}llrr@{}}
\toprule
CPUs & Forced plan & Original (s) & Revised (s) \\
\midrule
4 & Parallel & 119.97 & 82.96 \\
4 & Direct & 94.09 & 68.82 \\
16 & Direct & 94.54 & 67.33 \\
16 & Parallel & 106.13 & 71.84 \\
\midrule
\multicolumn{2}{@{}l}{Total worker time} & 414.73 & 290.95 \\
\bottomrule
\end{tabular}
\caption{Model-authored render-first orchestration on four matched development cases. Each pair uses the same forced physical plan. Original and revised runs are temporally separated.}
\label{tab:renderfirst}
\end{table}

\paragraph{Feedback changes plan selection.}
Whole-plan observations update a selector that can choose Direct or parallel execution. With the same revised source, feedback changes the twelve-clip landscape plan to Direct and leaves the mixed-orientation task unchanged (Table~\ref{tab:historicalfeedback}). Across eight suffix pairs, worker time falls from 227.85\,s to 186.70\,s, a reduction of 18.1\%. A subsequent eight-suffix control measures 178.14\,s for fixed Direct and 179.23\,s for feedback. Thus, feedback recovers an effective existing choice; it does not outperform that stronger fixed control. Output checks also identify audio residuals and mixed-orientation geometry errors.

\begin{table}[ht]
\centering
\setlength{\tabcolsep}{4.5pt}
\begin{tabular}{@{}lrrr@{}}
\toprule
Task & CPUs & \shortstack[r]{Prior\\(s)} & \shortstack[r]{Feedback\\(s)} \\
\midrule
Mixed, 6 clips & 4 & 23.22 & 23.48 \\
Mixed, 6 clips & 16 & 9.68 & 10.17 \\
Landscape, 12 clips & 4 & 60.32 & 43.26 \\
Landscape, 12 clips & 16 & 20.70 & 16.45 \\
\bottomrule
\end{tabular}
\caption{Mean suffix worker time with the same revised source, two ordered repetitions per cell on one learning trajectory. Both compositions request 24 seconds. Feedback changes the landscape plan and preserves the mixed-task plan.}
\label{tab:historicalfeedback}
\end{table}

\paragraph{Scheduling and overlap.}
Eight independent wait/admit pairs start from a common prediction state and use the same subsequent FIFO rule. Admit completes the stream sooner in all pairs, with mean saving 0.9\%, and brings completion of the queued task forward by 5.09\,s. The running task's remainder increases by 40.16\,s. This measures the trade-off between useful overlap and active-task contention.

\paragraph{Comparisons preceding Mac confirmation.}
The model selects three alternatives on the fixed task: two ranges, three ranges, and materialization followed by fusion. Their candidate/reference times are 15.31/14.21\,s, 13.98/14.46\,s, and 9.75/14.31\,s. All six outputs complete. These observations enter the successor proposal before the separate confirmations in Section~\ref{sec:performancecase}.

\section{Delivery-Recovery Diagnostic}
\label{app:recovery}
\label{sec:experientialcase}

A delivery-recovery comparison precedes the performance case. The initial fused $H_0$ produces 240/252/216 of the required 312 frames; $H_P$ completes. A feedback successor retains the failing action. Three fresh blocks with the initial policy and an additional-call no-feedback control give nine further failed outputs.

Both branches then receive the same clarified complete-delivery objective. One call receives the execution outcomes; the matched call omits those outcomes. The resulting $H_2$ chooses four clip-aware ranges with LPT and completes, while the no-feedback control remains incomplete (Table~\ref{tab:experientialcase}). $H_2$ has the reference's physical action and ratio 1.020. This comparison measures recovery of output validity. The performance case subsequently compares different complete-delivery plans.

\begin{table}[t]
\centering
\setlength{\tabcolsep}{5pt}
\begin{tabular}{@{}lrr@{}}
\toprule
Policy & Deliveries & Latency / $H_P$ \\
\midrule
No-feedback $H_{\mathrm{blind2}}$ & 0/3 & Undefined \\
Feedback $H_2$ & 3/3 & 1.020 \\
Reference $H_P$ & 3/3 & 1.000 \\
\bottomrule
\end{tabular}
\caption{Recovery on three repetitions of one task. $H_2$ restores delivery but uses the reference's physical action and has no speed advantage.}
\label{tab:experientialcase}
\end{table}

\section{Development Cost}
\label{app:cost}

The Mac performance case uses two model calls, 553,255 tokens, and six paired experiments, including the three fresh confirmations. API call time totals 64.8\,s. Incremental end-to-end study time is 11.6 minutes, including artifact transfer; post-timing output decoding takes 19.6\,s. The Linux allocation experiment uses 25 generations and 1,561,455 generation tokens. Its experimental runtime is 11.8 hours, including failed setup, followed by 1.8 hours of output decoding. These are study costs; inherited experience acquisition and monetary costs remain incomplete.

\texttt{EXPERIMENT\_SUMMARY.json}, included with the paper's source package, records the reported metrics, model identifiers, fixed-plan comparisons, failures, and known cost components. It retains unsuccessful setup, model waiting/loading, and infrastructure costs with their overlap relations.

\end{document}